\documentclass[10pt,twocolumn,letterpaper]{article}

\usepackage{times}
\usepackage{epsfig}
\usepackage{graphicx}
\usepackage{amsmath}
\usepackage{amssymb}

\usepackage{booktabs}
\usepackage{arydshln}
\usepackage{multirow}

\usepackage[breaklinks=true,bookmarks=false]{hyperref}

\begin{document}

\title{Conditioning Diffusion Models via Attributes and Semantic Masks for Face Generation}

\date{}

\author{Nico Giambi and Giuseppe Lisanti\\
Department of Computer Science and Engineering, University of Bologna\\
{\tt\small nico.giambi@studio.unibo.it, giuseppe.lisanti@unibo.it}
}

\maketitle

\begin{abstract}
Deep generative models have shown impressive results in generating realistic images of faces.
GANs managed to generate high-quality, high-fidelity images when conditioned on semantic masks, but they still lack the ability to diversify their output.
Diffusion models partially solve this problem and are able to generate diverse samples given the same condition. 
In this paper, we propose a multi-conditioning approach for diffusion models via cross-attention exploiting both attributes and semantic masks to generate high-quality and controllable face images.
We also studied the impact of applying perceptual-focused loss weighting into the latent space instead of the pixel space.
Our method extends the previous approaches by introducing conditioning on more than one set of features, guaranteeing a more fine-grained control over the generated face images. 
We evaluate our approach on the CelebA-HQ dataset, and we show that it can generate realistic and diverse samples while allowing for fine-grained control over multiple attributes and semantic regions. 
Additionally, we perform an ablation study to evaluate the impact of different conditioning strategies on the quality and diversity of the generated images.
\end{abstract}


\section{Introduction}
    Image synthesis has recently become a hot topic, mostly thanks to the vast number of successful applications proposed in the literature. 
    Among the different generation tasks, several works have focused the attention on semantic face image synthesis. Most of these solutions rely on GANs' and their ability to generate high-quality and high-fidelity results~\cite{CLADE, INADE, controlgan, DDGAN,DSCGAN}.
    However, their uni-modal nature prevents them to generate diverse samples~\cite{semantic}. 
    Diffusion Models~\cite{DDPM, LDM, semantic, P2, D_beat_G}
    have proven to compete with GANs in both quality and fidelity while being multi-modal generators.
    They are parameterized Markov chains that optimize the variational lower bound on the likelihood function to generate samples matching the data distribution.
    In order to generate an image, DMs iteratively refine a Gaussian noise via a Denoising process, that is implemented with a UNet~\cite{Unet} backbone.
    
    In this paper, we show how to achieve and surpass the actual state-of-the-art for semantic face image synthesis, following three main evaluation criteria: quality, fidelity, and diversity.
    In order to improve quality, we employ a reweighed loss function~\cite{P2} aimed to favor perceptual quality over unperceivable high-frequency details.
    A better fidelity is obtained by using a powerful conditioning mechanism, which in our case is cross-attention~\cite{crossattention}, combined with semantically and spatially rich encodings.
    Then, we examine diversity by leveraging Diffusion Models' natural ability to generate multi-modal images, using stricter/looser conditioning, resulting respectively in more consistent/diverse generated images.
    Finally, we propose a way to exploit cross-attention in order to condition a diffusion model with multiple features at once, allowing a higher degree of control over the generation process.
    In our case, we consider both facial attributes and semantic masks, but the same idea could be extended to any other domain and set of features.
    For example, it could be possible to condition a model with both a semantic layout and a certain time of the day in order to generate landscapes with the right colors and shading or combine sketches and textual descriptions in order to generate images of suspects in the forensics field.
    Our contributions can be summarized as follows:
    \begin{itemize}
        \item the analysis of perception prioritizing loss weighting~\cite{P2} in the latent space of Latent Diffusion Models~\cite{LDM}, which enhances the quality of generated samples without increasing the model's size or training/sampling time.
        \item a multi-conditioning solution to impose more strict and precise control over the generated images. 
        This mechanism lets the user combine spatial-only conditioning, like semantic masks, with descriptive features, like colors, shades, or level of detail from attributes. Additionally, we show that multi-conditioning causes a slight decrease in quality but results in high fidelity on both the provided conditioning.
        \item a state-of-the-art model for semantic face image synthesis, surpassing previous works in terms of generated images' quality, fidelity, and diversity.
    \end{itemize}

    \begin{figure*}
        \includegraphics[width=\textwidth]{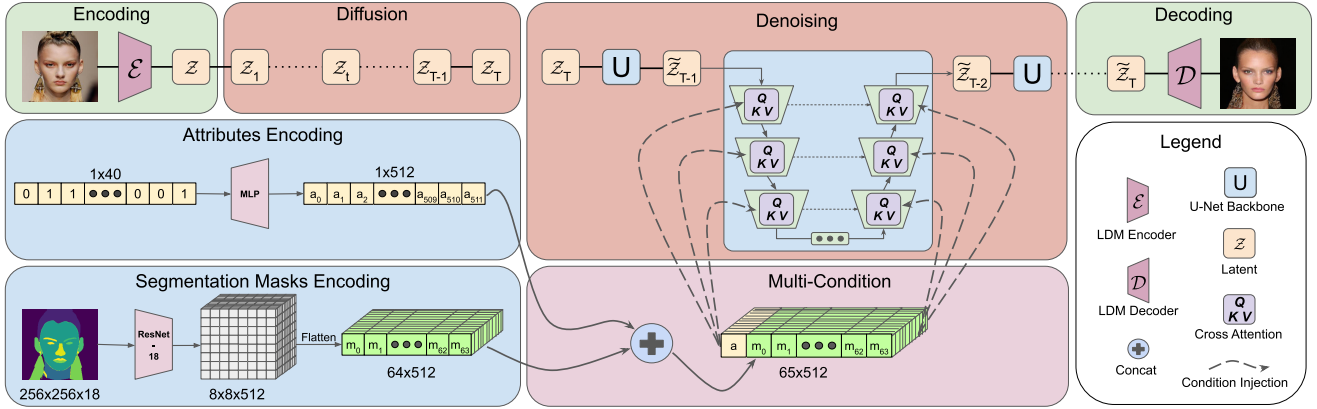}
        \caption{Multi condition model schema. Single conditioning and Unconditioned generation are simplifications of this model.}
        \label{fig:model}
    \end{figure*}
    
\section{Related Works}

In the following, we analyze some of the most recent works based on denoising diffusion models and solutions that generate face images from semantic masks or attributes.
    \subsection{Denoising Diffusion Models} 
        Recently, Diffusion Models (DM)~\cite{DM}, have achieved state-of-the-art results in various generative tasks, such as Image Synthesis~\cite{DDPM, D_beat_G, cascaded}, Image Inpainting~\cite{LDM} and Image-to-Image Translation~\cite{CycleDiffusion}.
        Ho~\emph{et~al.}~\cite{DDPM} performed an empirical analysis to propose a reweighted loss function. As an extension, Choi~\emph{et~al.}~\cite{P2} generalized this concept in order to establish a Perception Prioritized (P2) Weighting of the training objective.
        Recently, Rombach~\emph{et~al.}~\cite{LDM} obtained outstanding results by composing a Latent Diffusion Model (LDM) in order to compress data and denoise them in a smaller latent space, reducing by a great margin the amount of resources used for both the training and the sampling stages.
        Henry~\emph{et~al.}~\cite{CycleDiffusion} analyzed the latent variables of different implementations of DMs~\cite{DDIM, DDGAN, DiffAE} to perform Unpaired Image-to-Image Translation. 
        We leverage these solutions in order to train a model which is able to maximize the quality, fidelity, and diversity generation criteria.

    \subsection{Attributes Controlled Generation}
        Attributes Controlled Generation can both indicate synthesis and editing.
        In the last few years, \emph{attributes-controlled} image editing has received a lot of attention~\cite{attr_1_stargan, attr_2_high_fidelity, attr_3_guided_style, attr_4_residual, attr_5_relgan, attr_6_hierarchical}, while \emph{attributes-conditioned} image synthesis has not been of major interest.
        Li~\emph{et~al.}~\cite{stylet2i} proposed a text-to-image generation process that relies on the text transposition of the CelebA-HQ attributes and 
        compared their results with other similar studies~\cite{tedigan, controlgan, daegan}. 
        We will compare the performance of our model to these methods since they are the closest to our solution and provide quantitative results in terms of FID~\cite{FID}.
        Unlike previous approaches, our study focuses on attributes-conditioned image synthesis. 
        We train LDM on the complete set of 40 CelebA-HQ~\cite{celebahq} attributes and show its capability in  generating high-quality and high-fidelity samples. This level of control could potentially facilitate the development of solutions that can produce datasets for various tasks, such as Image-to-Image Translation or Attributes-Controlled Image Editing.

    \subsection{Semantic Image Synthesis} 
        
        Over the years, semantic image synthesis has been mainly addressed by exploiting GAN-based~\cite{GAN} models. GAN-based approaches like  Pix2PixHD~\cite{Pix2Pix}, SPADE~\cite{SPADE}, CLADE~\cite{CLADE}, SCGAN~\cite{SCGAN} and SEAN~\cite{SEAN} focus on generating unimodal images. 
        Other works like BycicleGAN~\cite{Bycicle}, DSCGAN~\cite{DSCGAN} and INADE~\cite{INADE} aim to explore multimodal generation, which consists in generating high-fidelity and diverse samples.
        Recently, diffusion models have proved to obtain generation results with higher  diversity and fidelity~\cite{LDM, semantic}. Wang~\emph{et~al.} proposed Semantic Diffusion Model (SDM)~\cite{semantic}, for semantic image synthesis through DMs. SDM processes the semantic layout and the noisy image separately, in particular, it feeds the noisy image to the encoder stage of the U-Net model and the semantic layout to the decoder, using multi-layer spatially-adaptive normalization operators. This results in higher quality and semantic correlation of the generated images.
        
        Differently from this approach, we exploit LDM's cross-attention~\cite{crossattention} mechanism to inject semantically relevant spatial features into multiple U-Net stages.
        Cross-attention allows more flexible and powerful control over the generation results, 
        enabling us to execute multi-conditioning of a DM by utilizing both semantic layouts and facial attributes.\\
        
\section{Proposed Method}

In this section, we first provide some details about latent diffusion models and the loss weighting exploited in our model~\cite{P2}. Then we illustrate how semantic masks and attributes can be used to condition the generation process.

    \subsection{Latent Diffusion Model}
        Rombach~\emph{et~al.} introduced Latent Diffusion Model (LDM)~\cite{LDM} to minimize DMs' computational demands while maximizing the generated samples' quality. They proposed a general purpose, perceptually focused Encoder $(\mathcal{E})$ in order to project the high-quality input image from pixel space to a lower dimensionality, semantically equivalent,  latent space. 
        The smaller input helps to speed up the training since it is possible to feed the model with bigger batches, but the most important advantage can be observed during the sampling. The iterative denoising process, indeed, usually requires about 500 steps. 
        Therefore, reducing the Gaussian Noise size by a factor of 4, on each spatial dimension, results in a much faster sampling in the DM's space.
        Additionally, both the Encoder and the Decoder only need a single pass, meaning they bring a negligible overhead to the denoising process computational cost. 
        This Encoding-Decoding process separates the \emph{semantic compression} and \emph{perceptual compression} phases.
            The first is completely handled by the Encoder-Decoder
        while the latter is managed through the U-Net backbone, which can use all its parameters to focus on the perceptual part of the denoising.
        Since LDM achieved outstanding results for various Unconditioned and Conditioned tasks, we decided to base our work on this particular framework.

    \subsection{Perception Prioritized Loss Weighting}
    Choi~\emph{et~al.}~\cite{P2} analyzed the performance of the different stages of the DMs denoising process.
    By using perceptual measures like LPIPS~\cite{LPIPS}, they separate the diffusion process in three stages, parametrized on a Signal-to-Noise Ratio (SNR)~\cite{VDM} depending on the variance schedule.
    These stages define when different levels of detail are lost during the diffusion, or vice-versa when they are generated in the denoising process.
    In the first stage of denoising, coarse details like color and shapes are generated. 
    Then, in the content stage, more distinguishable features come up.
    In the final stage, the fine-grained high-frequency details are refined and most of them are not perceivable by the human eyes. 
    
   To this end, they proposed a Perception Prioritized (P2) Weighting of DM's loss function:
    \begin{equation}
        \begin{aligned}
            L_{P2}^t 
            &= \frac{1}{(k + SNR(t))^\gamma} \mathbf{E}_{\mathbf{x}, \mathbf{\epsilon}} \Big[\| \epsilon - \epsilon_\theta(\mathbf{x}_t, t)\|^2\Big]
        \end{aligned}
        \label{eq:lambda_t_prime}
    \end{equation}
    where \textit{k} is a stabilizing factor that avoids exploding weights for small SNR values, usually set to 1, and $\gamma$ is an arbitrary exponent that gives more or less importance to the re-weighting.
    We decided to explore the possibility of employing this loss weighting in the latent space of LDM\footnote{For the detailed mathematical derivation, please refer to the supplementary material.}, instead of the pixel space as done in~\cite{P2}.
    This is achieved by modifying the original loss formulation of~\cite{LDM} as follows:
    \begin{equation}
        \begin{aligned}
            L_{LDM}^t 
            &= \mathbf{E}_{\mathcal{E}(x),y,\epsilon \sim \mathcal{N}(0,1), t} \Big[ \| \mathbf{\epsilon} - \mathbf{\epsilon}_\theta(\mathbf{z}_t, t, \tau_{\theta}(y))\|^2 \Big]
        \label{eq:loss_ldm}
        \end{aligned}
    \end{equation}
    by introducing the weighting factor from Eq.~\ref{eq:lambda_t_prime}:
    \begin{equation}
        \begin{aligned}
            L_{LDM}^t 
            &= \mathbf{E}_{\mathcal{E}(x),y,\epsilon \sim \mathcal{N}(0,1), t} \Big[ \frac{\| \mathbf{\epsilon} - \mathbf{\epsilon}_\theta(\mathbf{z}_t, t, \tau_{\theta}(y))\|^2}{(k + SNR(t))^\gamma} \Big].
        \label{eq:loss_ours}
        \end{aligned}
    \end{equation}
    In both Eq.~\ref{eq:loss_ldm} and Eq.~\ref{eq:loss_ours}, $z_t$ is the latent representation of the input image obtained by the Encoder $\mathcal{E}$ at diffusion timestep $t$, $\tau_{\theta}$ is the condition encoder model and $y$ is its input, which can be a segmentation mask, an attribute array, a text prompt or anything else.

    \subsection{Attributes and Mask Conditioning} \label{sec:crossattention}
    Conditioning a generative model consists in injecting some kind of information, such that the generated samples will reflect this property. 
    In GANs this information is usually injected exploiting a \emph{normalization layer}, like semantic region-adaptive normalization in SEAN~\cite{SEAN}, spatially conditioned normalization in SCGAN~\cite{SCGAN} and instance-adaptive denormalization in INADE~\cite{INADE}.
    DMs use a similar process to inject information into the denoising process. For example, Dhariwal and Nichol~\cite{D_beat_G} proposed the adaptive group normalization (AdaGN) to condition the DM on both the class embedding and the time-step after each group normalization layer, while Wang~\emph{et~al.}~\cite{semantic} proposed the multi-layer spatially-adaptive normalization in order to feed the segmentation masks into the decoder stage of the denoising U-Net.
    Rombach~\emph{et~al.}~\cite{LDM}, instead, exploited the spatial transformer~\cite{crossattention} as a flexible and powerful conditioning mechanism to be applied to a subset of layers of the U-Net.
    The spatial transformer is composed of three distinct components, the first of which is a self-attention mechanism, computed on the set of features from the relative U-Net layer.  
    The output of the self-attention is then summed to the input features via residual connection and provided as input to a cross-attention mechanism which combines information from the previous layer and the condition.
    The output is again summed to the input of the cross-attention and passed through an expansion-compression feed-forward neural network~\cite{crossattention} which provides the output, that represents the conditioned set of features.
    We decided to follow this approach for conditioning our model with:
    (i)  an encoding of binary attributes;
    (ii) an encoding of segmentation masks;
    (iii) a sequence obtained as the concatenation of the encoding from both attributes and segmentation masks (Fig.~\ref{fig:model}).
    As described above, among the different layers composing the spatial transformer, the cross-attention (CA) is the one responsible for the injection of the condition, and is defined as:    
    \begin{equation}
        \begin{aligned}
            CA(Q,K,V) = softmax \Big(\frac{QK^T}{\sqrt{d}}\Big) \cdot V, \\
            Q\in\mathbb{R}^{\phi \times (h \cdot d)},
            \hspace{10px}
            K\in\mathbb{R}^{\psi \times (h \cdot d)},
            \hspace{10px}
            V\in\mathbb{R}^{\psi \times (h \cdot d)},
        \end{aligned}
    \end{equation}
    where 
    $d$ is the dimension of each attention head output (i.e., $d=64$ as in~\cite{crossattention, LDM}), 
    $h$ is the number of attention heads,
    $K, V \in\mathbb{R}^{\psi \times (h \cdot d)}$ are computed from the encoded conditioning, 
    $Q\in\mathbb{R}^{\phi \times (h \cdot d)}$ is a representation obtained from the corresponding U-Net layer on which the spatial transformer is applied.
    The dimension $\phi$ results from flattening the U-net activations of the relative layer, while the dimension $\psi$ represents the length of the conditioning sequence.

    The final output of the conditioning $CA(Q,K,V)$ will have the same dimension as the initial input $Q\in \mathbb{R}^{\phi \times (d \cdot h)}$ and will be provided as \emph{conditioned} input to the next layer of the U-Net.
    It can be observed that the output shape doesn't depend on the conditioning sequence length $\psi$, and this allows us to provide a variable set of conditions. In particular, we evaluate three different conditioning: 
    \begin{itemize}
    \item the binary attributes conditioning which is obtained through an MLP that maps the 40 attributes
    to $\mathcal{Z}_{a} \in \mathbb{R}^{\psi_{a} \times (d \cdot h)}$ ;
        \item the mask conditioning, $\mathcal{Z}_{m} \in \mathbb{R}^{\psi_{m} \times (d \cdot h)}$, which is obtained by feeding the semantic mask to a ResNet-18;
        \item the multi-conditioning, $\mathcal{Z}_{mc} \in \mathbb{R}^{(\psi_{a} + \psi_{m}) \times (d \cdot h)}$, which is obtained by concatenating the two encodings along the $\psi$ axis;
    \end{itemize}

    For the last point, we decided to prune the ResNet-18 encoder up to just before the  Global Average Pooling layer, in order to keep more high-level semantic spatial information.
    Working with 256x256 images, our ResNet encoder maps the masks $m \in \mathbb{R}^{256 \times 256 \times 18}$ into
    $\mathcal{Z}_{m} \in \mathbb{R}^{(8 \cdot 8) \times (d \cdot h)}$. Our multi-condition encoder will then generate $\mathcal{Z}_{mc} \in \mathbb{R}^{(1 + 64) \times (d \cdot h)}$, one embedding for the attributes and $64$ for the flattened masks features, output of the ResNet-18.
    The whole pipeline with the conditioning mechanism is illustrated in Fig.~\ref{fig:model}.

     \begin{figure*}
     \centering
        \includegraphics[width=0.9\textwidth]{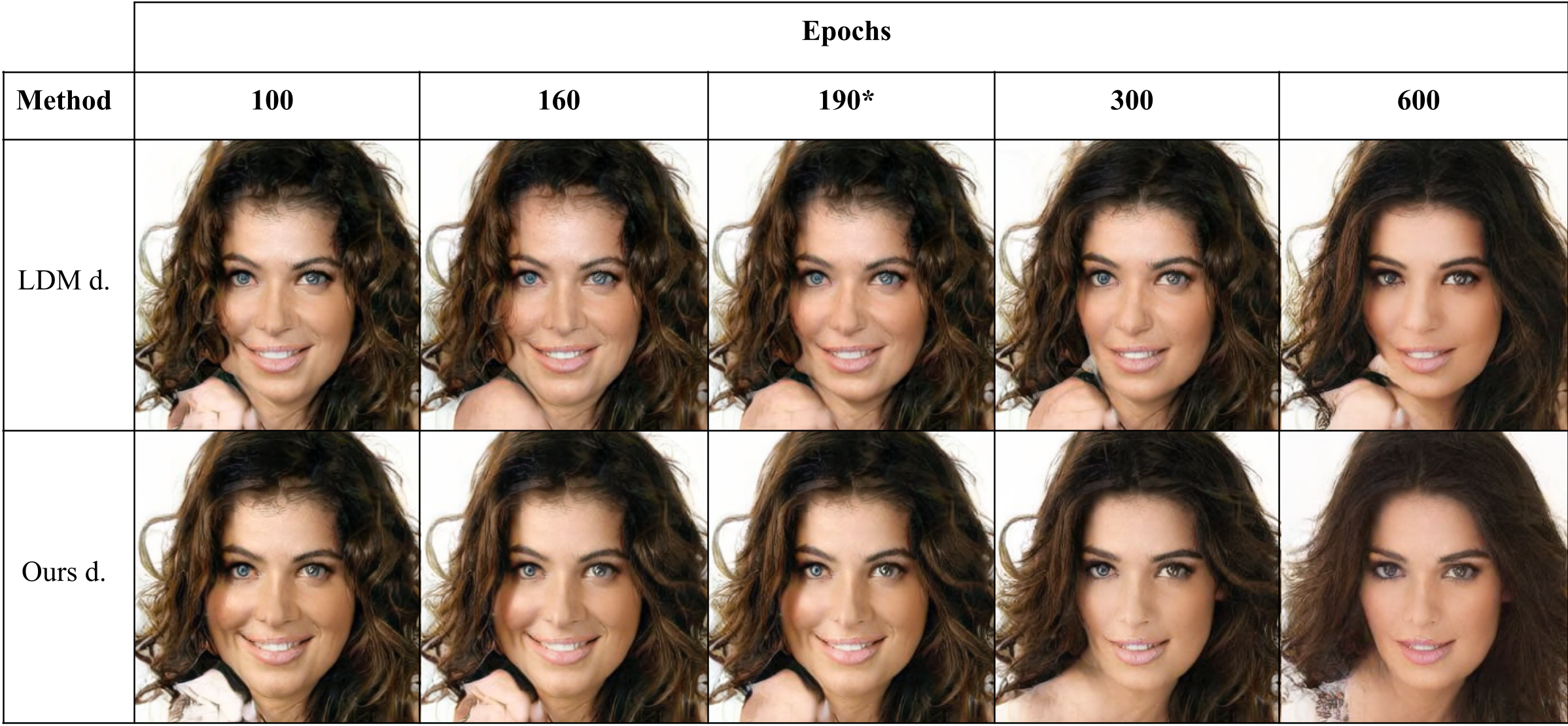}
        \caption{Unconditioned samples from LDM and the P2Weighted model at various training checkpoints. All the samples are generated from the same latent. * denotes the checkpoint used in the evaluation.}
        \label{fig:p2_vs_baseline}
    \end{figure*}

\section{Experimental results}
In the following, we first introduce the dataset and settings used in all our experiments. Then, we report both quantitative and qualitative generation results obtained by conditioning with attributes, semantic masks, or both.
All our models have been trained and tested using a single NVIDIA RTX 3090 with 24GB of memory.
\\
\textbf{Dataset and model}. 
All our experiments were performed on CelebAMask-HQ~\cite{celebamask} considering a resolution of 256x256 pixels. 
We use a train/validation split of 25.000/5.000, as in LDM~\cite{LDM}.
We use LDM's pre-trained encoder ($\mathcal{E}$) which maps images from the pixel space to a VQ-regularized latent space with a reduction factor of 4, hence performing diffusion and denoising on a 64x64 space. 
The latent space denoising U-Net ($\mathbf{U}$), the image decoder ($\mathcal{D}$), the attributes encoder ($\mathcal{E}_a$) and the ResNet-18 mask encoders ($\mathcal{E}_m, \mathcal{E}_{mnp}$) have all been trained from scratch.\\
\\
\textbf{Metrics}. 
We assess visual quality using Fréchet Inception Distance (FID)~\cite{FID} and Kernel Inception Distance (KID)~\cite{KID}.
For conditioned tasks, we also want to validate the correspondence between the generated samples and the condition, so we employ an accuracy score for masks, binary attributes, and multi-condition.
Moreover, we analyze a mean Intersection over Union (mIoU) of segmentation masks on mask-conditioned and multi-conditioned generation, more details in Sec.~\ref{sec:masks}.
We are also interested in evaluating diversity among samples conditioned on the same set of features. 
In this case, we use LPIPS~\cite{LPIPS} to evaluate our three conditioning methods. For each feature combination, we measure LPIPS among 10 samples. \\
        
In all our tables, when a number follows a metric's name, it means that all results shown in that table are computed on that specific  amount of samples. Otherwise, if a number is not specified, it means this information was not provided in the original paper. 
For unconditioned generation, we compute the metrics on 50K generated samples, while for conditioned generations we sample as many images as in the validation set (e.g, 5K samples), using the set of attributes or masks provided with the validation samples.
Each table includes metrics denoted by $\uparrow$ if higher is better, $\downarrow$ if lower is better.
All our samples are generated with 500 DDIM~\cite{DDIM} sampling steps.
We also denote our models using ``\textit{d.}'' if the results are taken from a deterministic sampling ($\eta$ = 0.0), or ``\textit{s.}'' if we used a stochastic sampling ($\eta$ = 1.0).

    \subsection{Unconditioned Image Synthesis}
        In this experiment, we want to analyze the improvement obtained by introducing P2 Weighting~\cite{P2} into LDMs.
        We train from scratch both the baseline LDM and the P2 weighted model using $\gamma = 0.5$ as suggested in~\cite{P2} for CelebA-HQ. 
        The two models have the exact same architecture and are both trained for 600 epochs, the only difference is in the objective function. 
        
        In Tab.~\ref{table:unc_epochs} we show the FID performance for different training checkpoints' on a subset of 10K generated samples.
        P2 improves the baseline FID at each checkpoint by 0.5 points, without increasing the model's number of parameters or its sampling time.    
        In Tab.~\ref{table:unc_scores}, instead, we report FID and KID results, also compared to previous works. We can observe that the proposed LDM, both with and without P2 weighting, obtains lower FIDs compared to most of the existing solutions.

        From now on we will employ the P2 weighting in all subsequent  experiments.
        In Fig.~\ref{fig:p2_vs_baseline} we compare qualitative examples generated from the same latent, using different checkpoints from both our P2-weighted model and the baseline LDM. 
        It is possible to appreciate that, after 100 epochs, our model has already reached satisfying generation stability while the baseline is still trying to converge.

    \begin{figure*}
        \centering
        \includegraphics[width=0.9\linewidth]{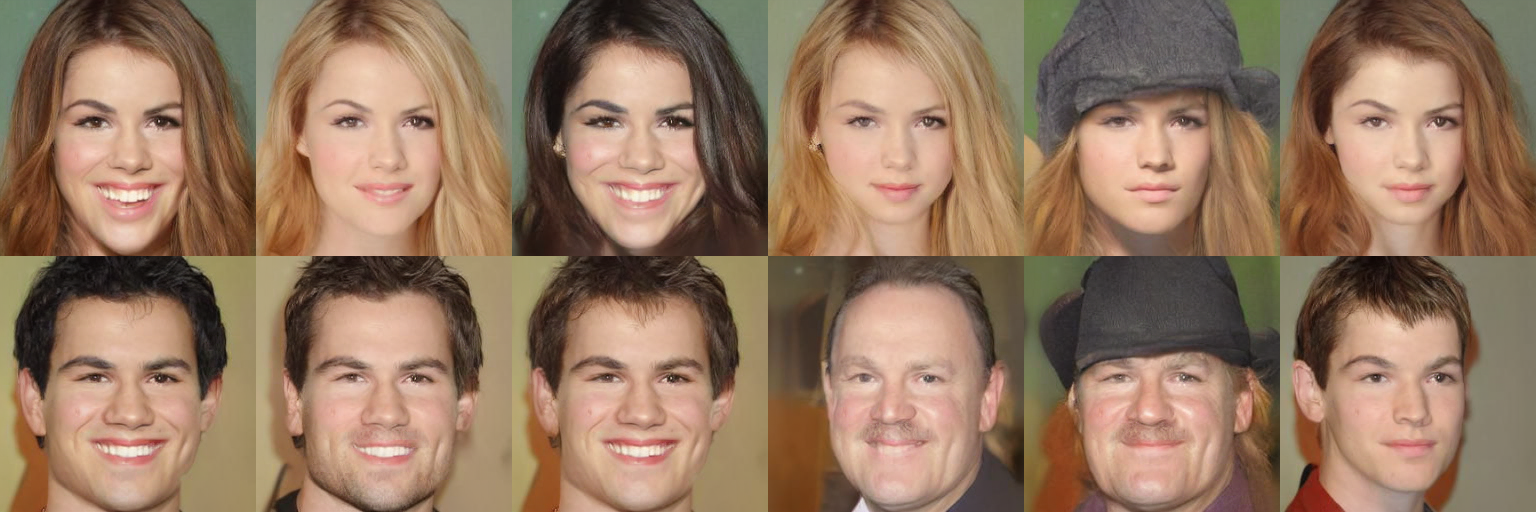}
        \caption{Generation examples obtained from the same latent (i.e., the same initial noise) using a deterministic DDIM. Each sample is conditioned on a random set of attributes chosen from the validation set.}
        \label{fig:attributes}
    \end{figure*}

        \begin{table}
            \begin{center}
                \begin{tabular}{l c c c c}
                    
                    \hline 
                    & \multicolumn{4}{c}{FID 10K $\downarrow$ at Epoch} \\
                    \cline{2-5}
                    \textbf{Method} & \textbf{100} & \textbf{160} & \textbf{190} & \textbf{300} \\
                    \hline
                    LDM d. & 7.79 & 6.97 & 6.97 & 8.52\\
                    Ours d. & \textbf{7.32} & \textbf{6.53} & \textbf{\underline{6.33}} & \textbf{8.14} \\
                    \hline
                \end{tabular}
            \end{center}
            \caption{FID 10K at various training epochs for LDM baseline and our P2 weighted LDM.}
            \label{table:unc_epochs}
        \end{table}

        \begin{table}
            \begin{center}
                \begin{tabular}{l c c }
                    \hline 
                    \textbf{Method} & \textbf{FID 50K}$\downarrow$ & \textbf{KID 50K}$\downarrow$ \\
                    \hline
                    PGGAN$^{\dag}$\cite{PGGAN} & 8.00 & - \\
                    DDGAN$^{*}$\cite{DDGAN} & 7.64 & - \\
                    LSGM$^{\dag}$\cite{LSGM} & 7.22 & - \\
                    UDM$^{\dag}$\cite{UDM} & 7.16 & - \\
                    WaveDiff$^{*}$\cite{WaveDiff} & 5.94 & - \\
                    LDM$^{\dag}$\cite{LDM} & 5.11 & - \\
                    StyleSwin$^{\ddag}$\cite{styleswin} & \textbf{3.25} & - \\
                    \hdashline
                    LDM d. & 5.88 & 0.0034 \\
                    LDM s. & 6.60 & 0.0036 \\
                    Ours d. & 5.42 & \textbf{0.0032} \\
                    Ours s. & 6.15 & 0.0033 \\
                    \hline
                \end{tabular}
            \end{center}
            \caption{Qualitative metrics computed on 50K samples.
            $*, \dag, \ddag$ means the corresponding result is taken from~\cite{WaveDiff},~\cite{LDM},~\cite{styleswin} respectively.}
            \label{table:unc_scores}
        \end{table}

    \subsection{Attributes Conditioned Synthesis} \label{sec:attributes}
    In this section, we'll show how a simple attributes encoding can successfully condition DMs via cross-attention, both quantitatively and qualitatively.
    We implemented our attributes encoder as a simple MLP which maps the set of 40 binary attributes into an embedding of dimension $d=512$.
    We feed this to the diffusion model via cross-attention, as detailed in Sec.~\ref{sec:crossattention}.
    We didn't find any significant previous work on this specific task, so we compare our results to StyleT2I~\cite{stylet2i} and other text-conditioned models on CelebA-HQ.
    In these solutions, the text is usually formed by composing phrases using keywords that correspond to the name of the binary attributes.
    
    From Tab.~\ref{table:attr_score} we can appreciate how the proposed conditioned model outperforms these solutions by a great margin, in terms of FID.
    In order to assess the conditioning fidelity of our model, we fine-tuned a ResNet-18 network on the CelebA-HQ training set to perform a multi-label attribute classification.
    The classifier obtains a \textbf{90.85\%} accuracy on the ground truth validation images, while the samples generated by our model (i.e., obtained by conditioning with the set of attributes from the validation set) obtain a classification accuracy of \textbf{90.53\%}, which confirms the capability of our model to generate samples which reflect the provided attributes. 
    
    \begin{table}
            \begin{center}
            \resizebox{\columnwidth}{!}{
                \begin{tabular}{l c c c c}
                    \hline 
                    \textbf{Method} & \textbf{FID $\downarrow$} & \textbf{KID $\downarrow$} & \textbf{Acc.(\%)$\uparrow$} & \textbf{LPIPS $\uparrow$} \\
                    ControlGAN\cite{controlgan} & 31.38 & - & - & -\\
                    DAE-GAN\cite{daegan} & 30.74 & -  & - & -\\
                    TediGAN-B~\cite{tedigan} & 15.46 & -  & - & -\\
                    StyleT2I\cite{stylet2i} & 17.46 & -  & - & -\\
                    \hdashline
                    Ours d. & \textbf{8.83} &  \textbf{0.0028} & 90.53 & -\\
                    Ours s. & 9.18 & \textbf{0.0028} & \textbf{91.14} & \textbf{0.549}\\
                    \hline
                \end{tabular}
                }
            \end{center}
            \caption{FID, KID, accuracy (Acc.), and LPIPS for attributes conditioned synthesis (bottom) and text conditioned synthesis (top).
            The text-conditioned results (top) are taken from StyleT2I~\cite{stylet2i}.}
            \label{table:attr_score}
        \end{table}

In Fig.~\ref{fig:attributes} we show some samples generated from the same noise. It could be observed that the output share a similar physiognomy, which differs just for the presence or absence of different attributes. This behavior was also observed in~\cite{CycleDiffusion}.

        \begin{figure*}
        \centering
            \includegraphics[width=0.9\linewidth]{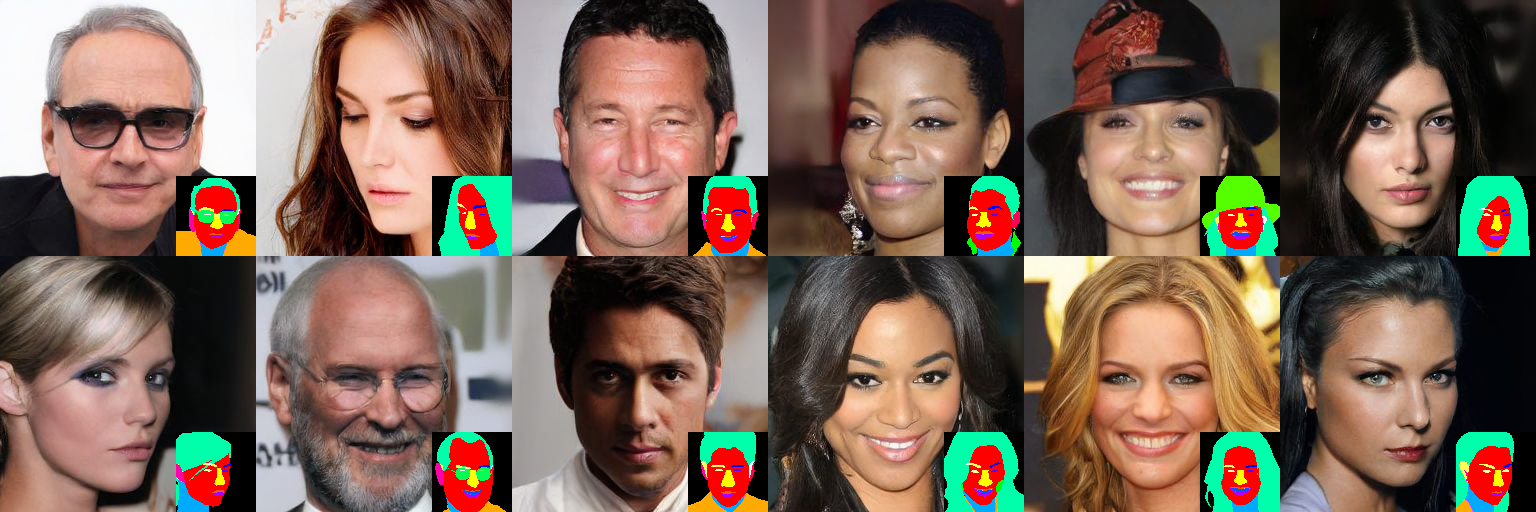}
            \caption{Qualitative samples generated from our model conditioned using $\mathcal{E}_{mnp}$ with their relative semantic masks.}
            \label{fig:masks_qualitative}
        \end{figure*}

    \subsection{Semantic Image Synthesis} \label{sec:masks}
        As for the attributes conditioned synthesis discussed in Sec.~\ref{sec:attributes}, we employ cross-attention to inject semantic information into our model. 
        This time, the encoder backbone is a pruned ResNet-18, with 18 input channels representing binary masks, one for each available part of the face, background excluded. 
        We tested two different conditions depending on the layers of the ResNet-18 encoder chosen to extract the features. 
        The first version, $\mathcal{E}_m$, is the full ResNet-18 backbone except for the classification layer, while the second, $\mathcal{E}_{mnp}$, also discards the Global Average Pooling layer, in order to preserve spatially relevant semantic information. The corresponding latent encodings are $\mathcal{Z}_m \in \mathbb{R}^{1 \times 1 \times 512}$ and $\mathcal{Z}_{mnp} \in \mathbb{R}^{8 \times 8 \times 512}$, which differ just for the spatial size.
        In Tab.~\ref{table:masks_score} we can see how both FID and conditioning fidelity are higher when $\mathcal{E}_{mnp}$ is employed, which demonstrates the capability of the cross-attention mechanism to leverage the information provided by the larger number of embeddings. 
        Both our conditioning methods outperform previous works, in terms of FID.
        Accuracy and mIoU are instead computed using off-the-shelf segmentation models\footnote{Source code available at: https://github.com/zllrunning/face-parsing.PyTorch}, by parsing semantic masks from our generated images and comparing them to their relative ground truth masks on which they were originally conditioned.  
        In Fig.~\ref{fig:masks_qualitative} we show some samples\footnote{More qualitative results and experiments can be found in the supplementary materials.} conditioned with $\mathcal{E}_{mnp}$. 
        Non-centered faces, glasses and hats don't pose any problems.

    \begin{table}
        \begin{center}
            \resizebox{\columnwidth}{!}{
                \begin{tabular}{l c c c c}
                    \hline 
                    \textbf{Method} & \textbf{FID $\downarrow$} & \textbf{Acc. (\%) $\uparrow$}  & \textbf{mIoU (\%) $\uparrow$} & \textbf{LPIPS $\uparrow$} \\
                    Pix2PixHD$^{\dag}$\cite{Pix2Pix} & 23.69 & 95.76 & 76.12 & -- \\
                    SPADE$^{\dag}$\cite{SPADE} & 22.43 & \textbf{95.93} & 77.01 & -- \\
                    SEAN$^{\dag}$\cite{SEAN} & 17.66 & 95.69 & 75.69 & -- \\                    GroupDNet$^{*}$\cite{GroupDNet} & 25.90 & -- & 76.10 & 0.365 \\
                    INADE$^{*}$\cite{INADE} & 21.50 & -- & 74.10 & 0.415 \\
                    SDM$^{*}$\cite{semantic} & 18.80 & -- & 77.00 & 0.422 \\
                    \hdashline
                    $Ours_{m}$ s.  & 8.41 & 91.52 & 75.80 & \textbf{0.469} \\
                    $Ours_{mnp}$ s. & \textbf{8.31} & 93.91 & \textbf{79.06}  & 0.446\\
                    \hline
                    Ground Truth & 0.0 & 95.51 & 81.79 & --\\ 
                    \hline
                \end{tabular}
                }
            \end{center}
            \caption{FID, accuracy (Acc.), mean Intersection over Union (mIoU) and LPIPS for masks conditioned synthesis. $\dag$/* denotes results taken respectively from SEAN~\cite{SEAN} and SDM~\cite{semantic}. Ground-Truth refers to masks parsed from the original validation set.}
            \label{table:masks_score}
        \end{table}

\begin{table*}
            \begin{center}
                \begin{tabular}{l c c c c c c}
                    \hline 
                    \textbf{Condition Encoder} & \textbf{FID $\downarrow$} & \textbf{KID $\downarrow$} & \textbf{Attr. Acc. (\%) $\uparrow$}  & \textbf{Masks Acc. (\%) $\uparrow$}  & \textbf{mIoU (\%) $\uparrow$} & \textbf{LPIPS $\uparrow$} \\ 
                    $\mathcal{E}_{a}$ d.  & 8.33 & 0.0028 & 90.53 & -- & -- & -- \\
                    $\mathcal{E}_{a}$ s.  & 9.18 & 0.0028 & \textbf{91.14} & -- & -- & \textbf{0.549} \\
                    \hdashline
                    $\mathcal{E}_{m}$ d.  & 8.49 & 0.0024 & -- & 91.36 & 75.14 & -- \\
                    $\mathcal{E}_{m}$ s.  & 8.41 & 0.0023 & -- & 91.52 & 75.80 & 0.469 \\
                    $\mathcal{E}_{mnp}$ d. & 8.43 & 0.0025 & -- & 93.77 & 78.67 & -- \\
                    $\mathcal{E}_{mnp}$ s. & \textbf{8.31} & \textbf{0.0021} & -- & 93.91 & 79.06 & 0.446 \\
                    \hdashline
                    $\mathcal{E}_{mc}$ d. & 8.39 & 0.0024 & 90.27 & 93.90 & 78.68 & -- \\
                    $\mathcal{E}_{mc}$ s.  & 8.39 & 0.0022 & 90.19 & \textbf{94.06} & \textbf{79.20} & 0.432 \\
                    \hline
                    Ground Truth & 0.0 & 0.0 & 90.85 &  95.51 & 81.79 & --\\ 
                    \hline
                \end{tabular}
            \end{center}
            \caption{Comparison across various metrics for different configurations of our model. FID and KID metrics are for sample quality, Acc and mIoU are for correspondence, and LPIPS for diversity.
            All the metrics are evaluated on 5K samples against their respective 5K images from the validation set, except for LPIPS which is computed on sets of 10 images for each of the 5K validation images and features. 
            \emph{(top)} attributes conditioning. \emph{(middle)} masks conditioning. \emph{(bottom)} multi-conditioning.}
            \label{table:multi_score}
        \end{table*}

        To analyze our model's ability to adapt to noisy masks, a second experiment has been conducted in which we (a) employ a face parsing model to extract the segmentation masks from the validation set (instead of extracting the mask from the generated image as in the previous experiment); (b) use these masks to condition our model (instead of using the ground-truth validation masks); (c) generate 5K samples on the new imperfect masks.
        In the last row of Tab.~\ref{table:masks_score} we show the accuracy and mIoU for the generated masks.
        We also computed FID for the 5K images obtained by conditioning our model with the noisy masks. The  FID obtained for this experiment, \textbf{8.20}, is lower than the one obtained with the default masks, indicating a good ability of our model to adapt to imperfect masks.

        We then performed a diversity study using LPIPS~\cite{LPIPS} as metric. We generated 10 samples for each segmentation mask in the validation set and computed an intra-class diversity score for each class. We report the average LPIPS results compared to previous works in Tab.~\ref{table:masks_score}.
        We decided to compute LPIPS only using stochastic samplers because of the greater differences which could show up in the samples due to the variance and hence more complex latent.
        As we can see, we surpass the previous models, based both on GANs and Diffusion Models, on quality, and diversity of the generated images while as regards fidelity we observe a slightly lower performance in terms of accuracy and a higher result in terms of mIoU.
        
        It is worth highlighting the fact that fidelity and diversity show an inverse behavior depending on the degree of conditioning we apply to our model.
        On one hand, the model conditioned with $\mathcal{E}_m$, uses only $1/64^{th}$ of the embedding compared to $\mathcal{E}_{mnp}$, which results in a less accurate encoding for semantic masks. 
        This reflects in a higher LPIPS and lower fidelity, expressed by both accuracy and mIoU. 
        On the other hand, using more spatially relevant conditioning allows for improving the results in terms of fidelity while observing a reduction in the capability of the model to diversify the generated images.

    \subsection{Multi Condition Image Synthesis}

\begin{figure*}
        \centering
            \includegraphics[width=0.9\linewidth]{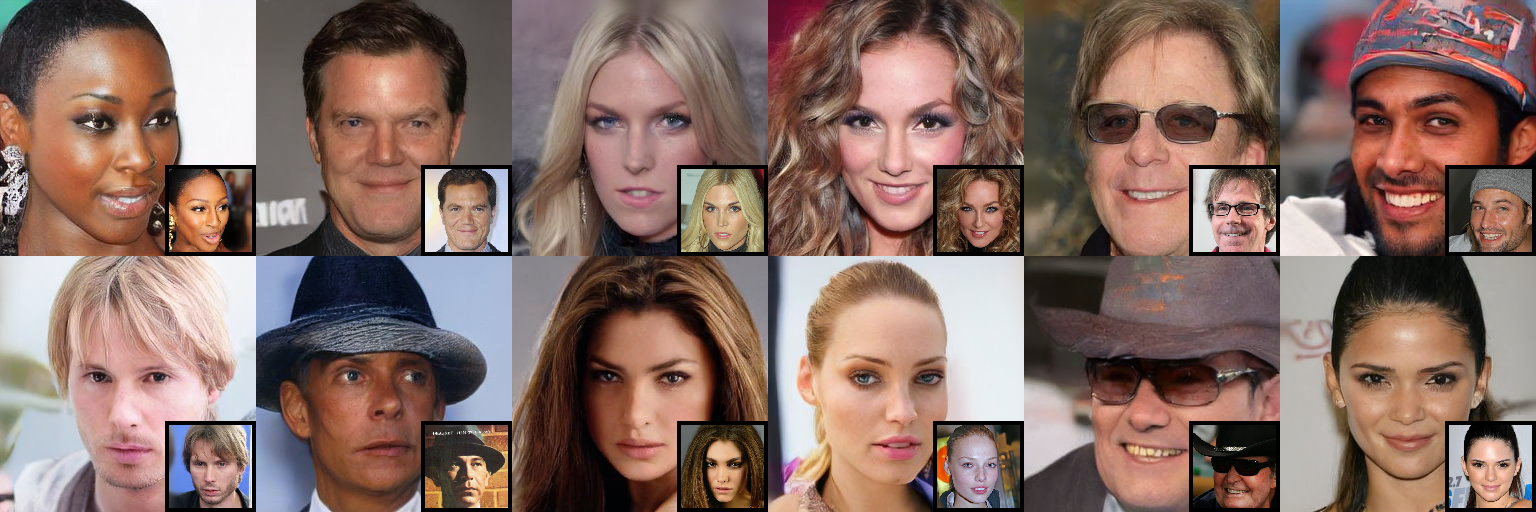}
            \caption{Qualitative samples generated from our model conditioned using both attributes and masks $\mathcal{E}_{mc}$ with, in the bottom-right, the validation set images from which the segmentation mask and attributes have been taken.}
            \label{fig:multi_qualitative}
        \end{figure*}

    \begin{figure*}
    \centering
            \includegraphics[width=0.95\linewidth]{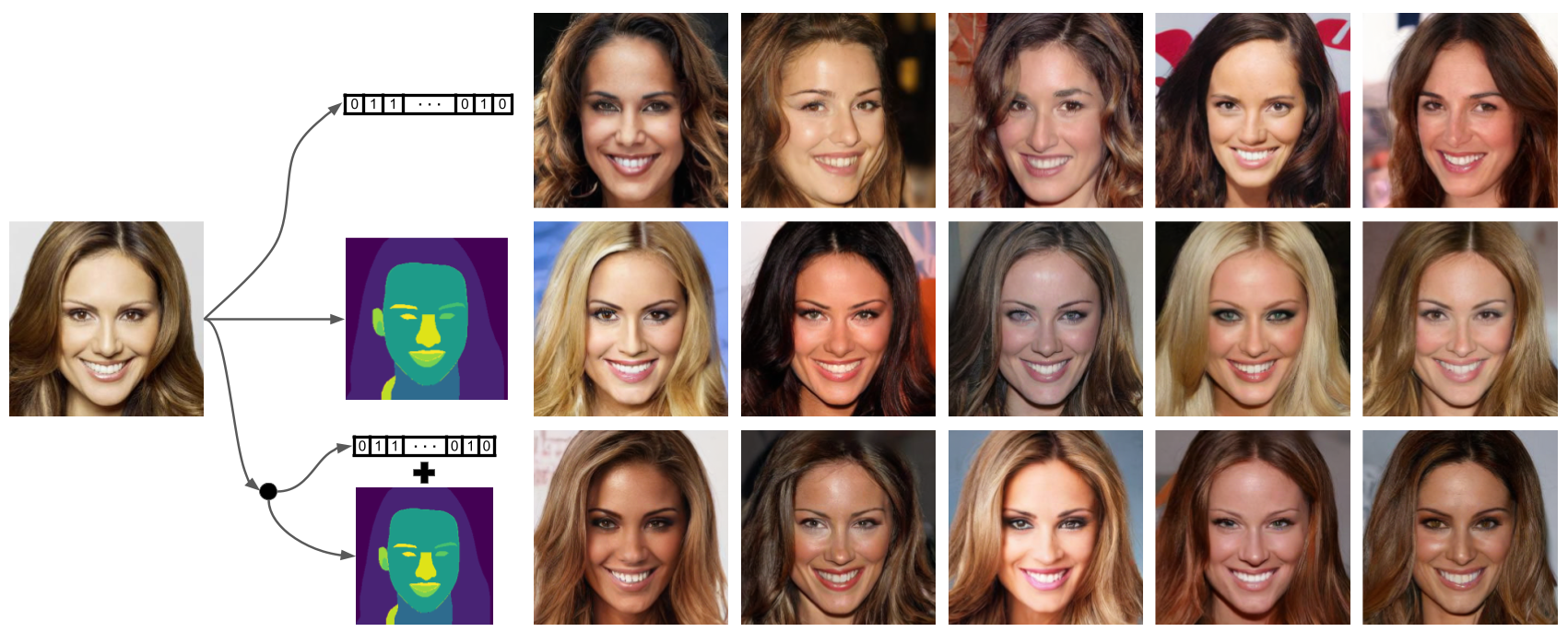}
            \caption{Qualitative samples showing the ability of our model to diversify its generated samples. 
            \emph{(left)} the reference image from the validation set.
            \emph{(top row)} the images generated when conditioning our model on the reference image's attributes.
            \emph{(central row)} the images generated when conditioning our model on the reference image's semantic mask.
            \emph{(bottom row)} the results obtained with our multi-condition encoder, using both attributes and semantic masks.}
            \label{fig:lpips}
        \end{figure*}

        As explained in Sec.~\ref{sec:crossattention}, we can exploit a property of cross-attention to inject two or more different sets of feature embeddings into any model via concatenation, before providing them as a condition into the spatial transformer.
        In particular, we combine CelebA-HQ's attributes and segmentation masks, to obtain even more fine-grained conditioning. 

        In our experiments we combined the attributes embedding, $\mathcal{Z}_{a} \in \mathbb{R}^{1 \times 512}$, and the flattened version of the mask embedding, $\mathcal{Z}_{mnp} \in \mathbb{R}^{(8 \cdot 8) \times 512}$.
        This results in a multi-condition embedding $\mathcal{Z}_{mc} \in \mathbb{R}^{65 \times 512}$ obtained via concatenation.

        In Tab.~\ref{table:multi_score} we report the results obtained using the multi-conditioned model against the attributes-conditioned and the mask-conditioned models. 
        It is worth noting that, the high fidelity observed on both attributes and masks results in lower FID and LPIPS, compared to single-conditioned models.
        
        Fig.~\ref{fig:multi_qualitative} shows some multi-conditioned examples generated by exploiting the segmentation masks and attributes of a face from the validation set, shown in the bottom-right of the generated samples.
        Finally, in Fig.~\ref{fig:lpips} we show the results obtained with the three different models, starting from the same attributes and mask.

\section{Conclusion}
    In this paper, we introduce a solution for face generation using diffusion models conditioned by both attributes and masks. We re-weight the loss terms of an LDM in a perception-prioritized fashion in order to achieve a higher quality of the generated samples.
    Then we explore the conditioned generation, first using attributes and then segmentation masks.
    We introduce a novel way to multi-condition a generative model exploiting cross-attention by joining the two conditions (i.e. attributes and semantic masks).
    Lastly, we evaluate both our single-conditioned and multi-conditioned models on a various range of metrics to assess quality, fidelity and diversity on CelebA-HQ~\cite{celebahq,celebamask} in terms of FID, KID, Accuracy, mIoU and LPIPS on three different types of conditioned generation.
    
    In our future work, we plan to explore the feasibility of implementing multiple conditions across various domains. We also intend to investigate and analyze more efficient techniques for encoding these different conditions.


{\small
\bibliographystyle{ieee_fullname}
\bibliography{egbib}
}

\onecolumn
\setcounter{section}{0}

\begin{center}
      \Large \textbf{Supplementary material for the paper ``Conditioning Diffusion Models via Attributes and Semantic Masks for Face Generation''}
\end{center}

\section{Latent P2 Weighting}

   In this section, we provide a detailed derivation of the P2 weighting and how we introduce it in our method.
  DMs could be seen as a particular kind of Variational Autoencoder (VAE), which can be trained by optimizing a variational lower bound (VLB), $L_{vlb} = \sum_t{L_t}$.
        For each time step $t$, the loss function could be defined as: 
        \begin{equation}
            \begin{aligned}
                L_t 
                &= \mathbf{E}_{\mathbf{x}, \mathbf{\epsilon}} \Big[\frac{\beta_t}{2 \alpha_t (1 - \bar{\alpha}_t)} \| \epsilon - \epsilon_\theta(\mathbf{x}_t, t)\|^2 \Big],
            \end{aligned}
        \end{equation}  
\noindent
    where $\alpha_t$, $\bar{\alpha}_t$ and $\beta_t$ represent the variance schedule, $\epsilon$ is the target Gaussian noise and $\epsilon_{\theta}$ is the parametrized U-Net model~\cite{DDPM}.

    When Ho~\emph{et~al.} proposed DDPM~\cite{DDPM}, they noticed that by removing the variance schedule-dependant coefficient, they obtained much better results and more stability at training time.
    Hence, they suggested using the following:
    \begin{equation}
        \begin{aligned}
            L_{simple}^t 
            &= \mathbf{E}_{\mathbf{x}, \mathbf{\epsilon}} \Big[ \| \mathbf{\epsilon} - \mathbf{\epsilon}_\theta(\mathbf{x}_t, t)\|^2 \Big] .
        \end{aligned}
    \end{equation}
    By removing the coefficient, the loss function is basically reweighted relative to the timestep term \textit{t}, as we can see here:
    \begin{equation}
        \begin{aligned}
            L_{simple}^t = \lambda_t L_t , \hspace{30px}             \lambda_t = \frac{2 \alpha_t (1 - \bar{\alpha}_t)}{\beta_t} 
        \end{aligned}
        \label{eq:lambda_t}
    \end{equation}

    The reason why this kind of reweighting works is explained by Choi~\emph{et~al.}~\cite{P2}. 
    They perform a broad analysis across different datasets, architectures and variance schedules in order to understand why the $ L_{simple}$ objective improved the perceived quality of the samples.
    By using perceptual measures like LPIPS~\cite{LPIPS}, they separate the diffusion process into three stages, parametrized on a Signal-to-Noise Ratio (SNR)~\cite{VDM} depending on the variance schedule.
    These stages define when different levels of detail are lost during the diffusion, or vice-versa when they are generated in the denoising process.
    In the first stage of denoising, coarse details like color schemes and shapes are generated. 
    Then, in the content stage, more distinguishable features come up.
    In the final stage, the fine-grained high-frequency details are refined and most of them are not perceivable by the human eyes. 
   They propose a Perception Prioritized (P2) Weighting of DM's Loss function:
    \begin{equation}
        \begin{aligned}
            L_{P2}^t 
            &=  \lambda_t^{'} L_t , \hspace{30px} \lambda_t^{'} = \frac{\lambda_t}{(k + SNR(t))^\gamma}
        \end{aligned}
        \label{eq:lambda_t_prime}
    \end{equation}
    \noindent
    where $\lambda_t$ is defined in Eq.~\eqref{eq:lambda_t}, \textit{k} is a stabilizing factor to avoid exploding weights for small SNR values, usually set to 1, and $\gamma$ is an arbitrary exponent to give more or less importance to the reweighting.
    P2 is a generalization of the $L_{simple}$ re-weighting, defined as follows:
    \begin{equation}
        Let \hspace{5px} \gamma = 0; \hspace{30px}
        \lambda_t^{'} = \frac{\lambda_t}{(k + SNR(t))^\gamma}\\
        = \lambda_t;        
    \end{equation}

    By increasing the value of $\gamma$ 
    the weights shift towards the coarse and content phases, representing the earlier stages of the denoising process, giving less and less importance to the loss terms corresponding to fine-grained unperceivable details.\\
    We decided to test P2 in the latent space of LDM since no previous work reports it.
    Both techniques seem to bring great improvement to DMs and don't show apparent conflicts when combined.
    We chose to use the proposed $\gamma$ values for the pixel-space dataset and analyze the experimental results. 
    We then consider the default conditioned LDM loss function:
    \begin{equation}
        \begin{aligned}
            L_{LDM}^t 
            &= \mathbf{E}_{\mathcal{E}(x),y,\epsilon \sim \mathcal{N}(0,1), t} \Big[ \| \mathbf{\epsilon} - \mathbf{\epsilon}_\theta(\mathbf{z}_t, t, \tau_{\theta}(y))\|^2 \Big]
        \end{aligned}
    \end{equation}
\noindent
    where $z_t$ is the latent representation of the input image obtained by the Encoder $\mathcal{E}$ at diffusion timestep $t$, $\tau_{\theta}$ is the condition encoder model and $y$ is its input, which can be a segmentation mask, an attribute array, a text prompt or anything else.
    We then updated the objective by adding the P2 weighting term:
    \begin{equation}
        \begin{aligned}
            L_{LDM}^t 
            &= \mathbf{E}_{\mathcal{E}(x),y,\epsilon \sim \mathcal{N}(0,1), t} \Big[ \frac{\| \mathbf{\epsilon} - \mathbf{\epsilon}_\theta(\mathbf{z}_t, t, \tau_{\theta}(y))\|^2}{(k + SNR(t))^\gamma} \Big]
        \end{aligned}
    \end{equation}
\noindent
    where the weight is defined in Eq.~\eqref{eq:lambda_t_prime}.

\section{Swapping components between masks}
\begin{figure*}[h]
        \includegraphics[width=0.92\linewidth]{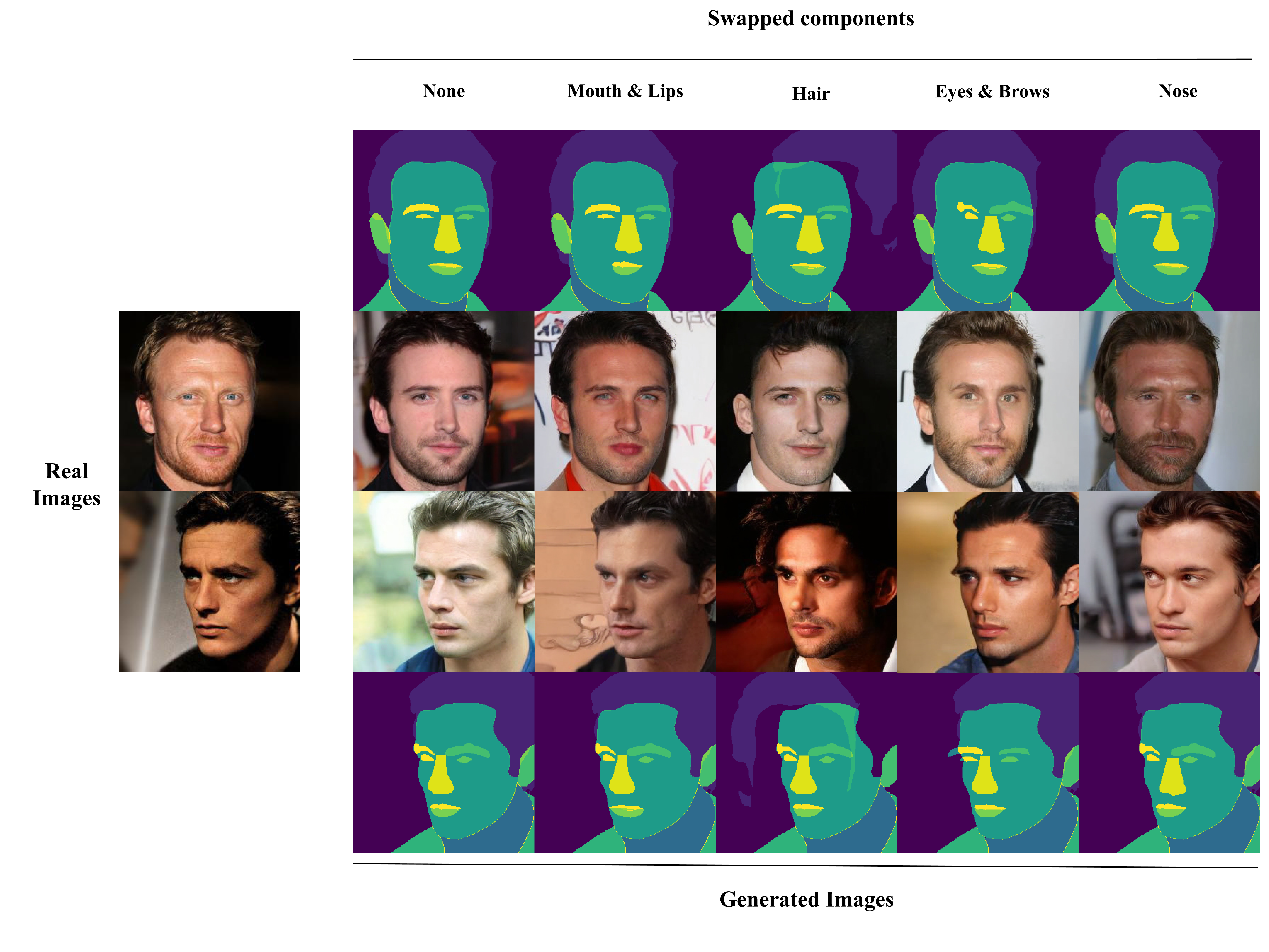}
        \caption{Samples generated by conditioning on incoherent segmentation masks, resulting from mixing differently oriented masks' components. 
        On the left, we show the real images from the validation set from which we took the masks. On the right, we show the swapped masks on top and bottom, and their relative generated samples just above or below.}
        \label{fig:scrambled}
    \end{figure*}

     To furtherly explore our model's ability to adapt to strange or incoherent masks, we tried swapping some components (i.e., mask channels) between pairs of segmentation masks, and used the resulting mixed mask as conditioning.
     In Fig.~\ref{fig:scrambled} we can see how our model can generate samples with high correspondence to the mask while trying to correct components that are no longer coherent with the rest of the mask.
     
     We choose to use this particular combination of faces since the components swapping is performed between two differently oriented faces. 
     As we can see by looking at the segmentation masks in Fig.~\ref{fig:scrambled}, we didn't perform any pre-processing on the mixed masks, however, the model is able to deal automatically with the misalignments.

\section{Failure Cases}
\begin{figure*}[h]
    \centering
        \includegraphics[width=0.999\linewidth]{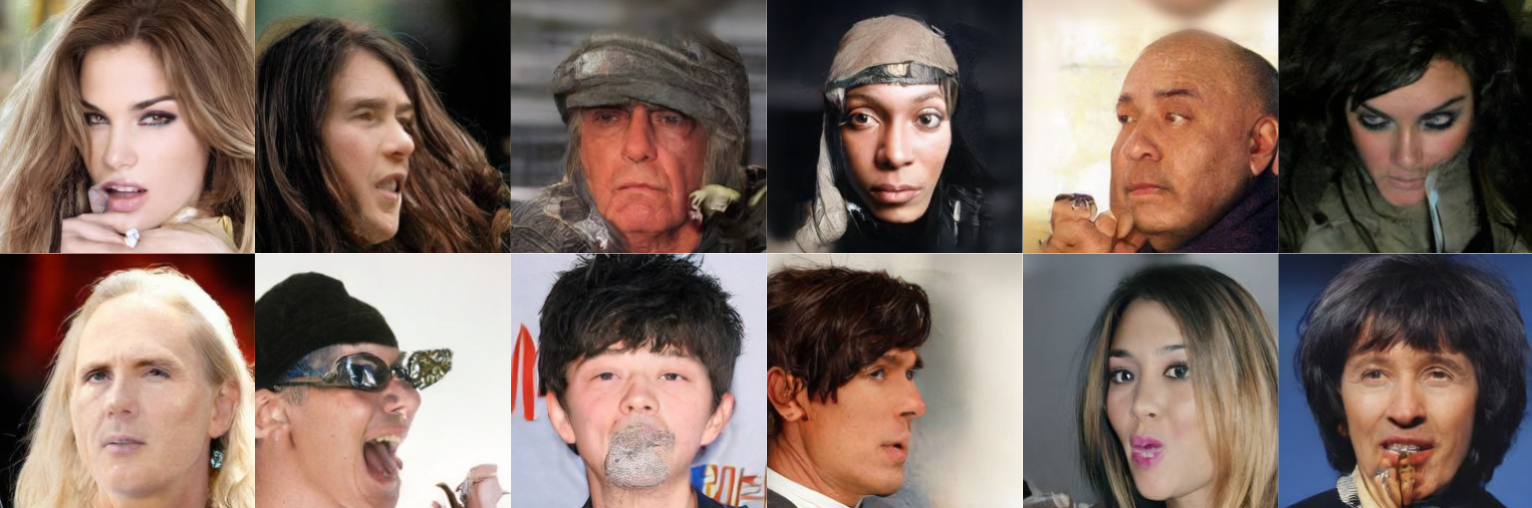}
        \caption{Worst handpicked samples generated by our mask-conditioned model.}
        \label{fig:bad}
    \end{figure*}

\begin{figure*}[h]
    \centering
        \includegraphics[width=0.999\linewidth]{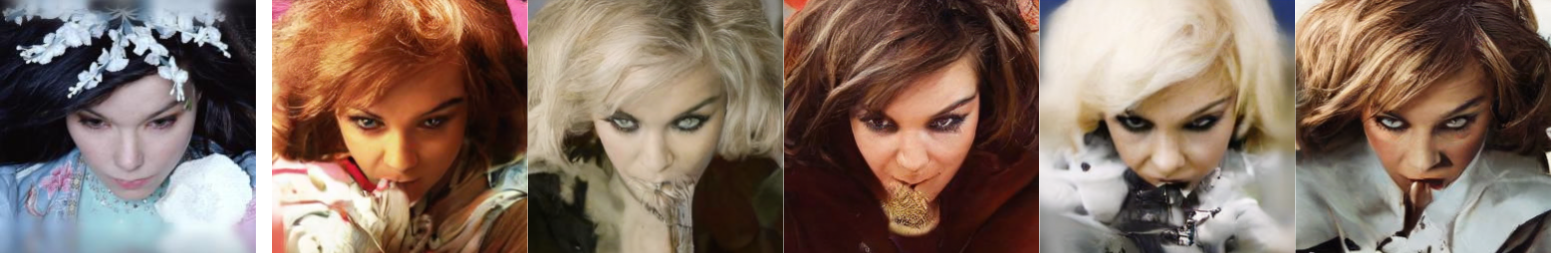}
        \caption{Failure cases generated by our mask-conditioned model on a peculiar mask.
        On the left, there's the original image from the validation set, while on the right we show our samples generated while conditioning on the reference image's semantic mask.}
        \label{fig:bad_mask}
    \end{figure*}

    In this section, we want to report some of our failure cases, represented by non-realistic images. 
    By looking through the generated samples, we noticed our unconditioned model rarely outputs unrealistic samples.
    Our conditioned models, though, sometimes produce bad samples, as shown in Fig.~\ref{fig:bad}. 
    This is a rare behavior since the faces in Fig.~\ref{fig:bad} are the only unrealistic results we were able to find among the 5K generated samples obtained using the segmentation masks of the validation set.
    By generating more samples conditioned on the same masks as the results reported in  Fig.~\ref{fig:bad}, we noticed there are two possible behaviors:  
 \emph{(1)} bad samples come from very peculiar masks, which are under-represented in the dataset, hence not reflecting the facial statistics learned by the model (see Fig.~\ref{fig:bad_mask});
 \emph{(2)} bad samples don't depend on bad segmentation masks but on a specific combination of mask and noise, where the conditioning strongly collides with the direction the noise is guiding towards, resulting in an unrealistic face. The noise indeed is relevant to the generated images since diffusion models tend to converge to similar latent spaces if they have the same variance schedule, as explained in~\cite{CycleDiffusion} and shown in Fig.~\ref{fig:attr}.
    
    It is also worth noting that we didn't find any major case of non-faithful images, with respect to attributes and/or masks, through the thousands of generated samples.
    Our model tends to prioritize the conditioning injection to the image's quality, resulting in faithful but unrealistic generated samples.
    Quantitatively, this behavior is described by all the results for the conditioned tasks, reported in the main manuscript, where a high-fidelity batch of generated samples brings an increase in FID.
    Our multi-conditioned scenario fits in this behavior since it performed slightly worse in terms of FID than the single-condition models, but reached high fidelity for both conditionings.

\section{Supplementary Qualitative Results}

\begin{figure}[!h]
    \centering
        \includegraphics[width=0.9\linewidth]{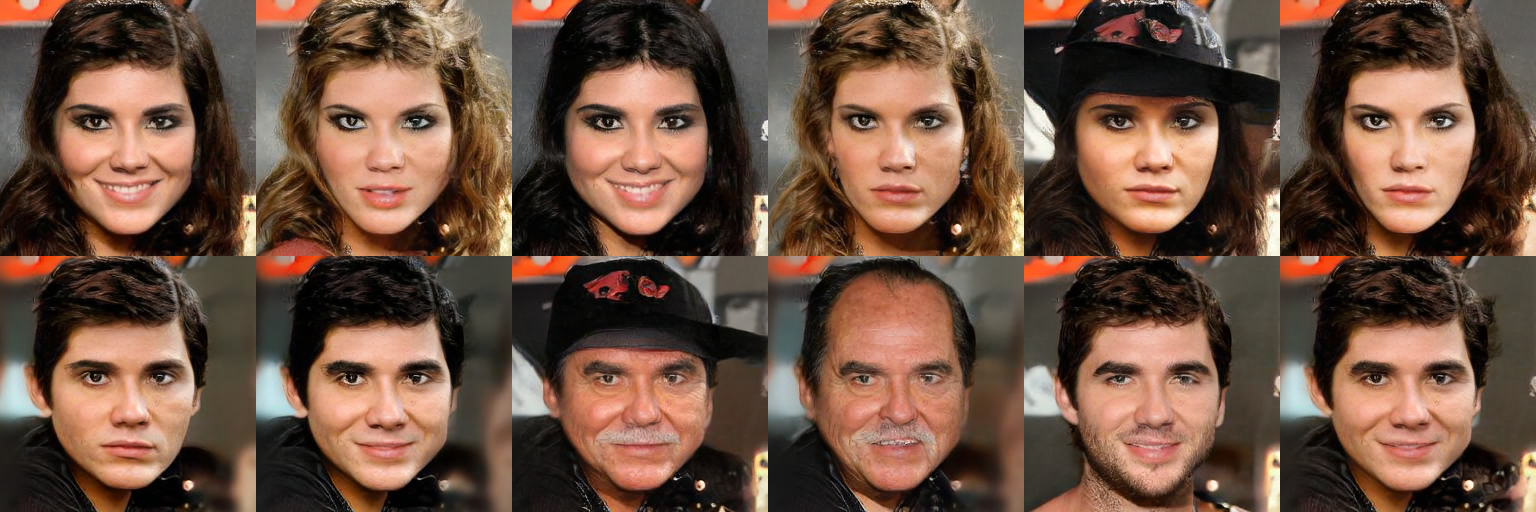}
        \caption{Additional results for Attributes-Conditioned Generation.}
        \label{fig:attr}
    \end{figure}

\begin{figure}[!h]
    \centering
        \includegraphics[width=0.9\linewidth]{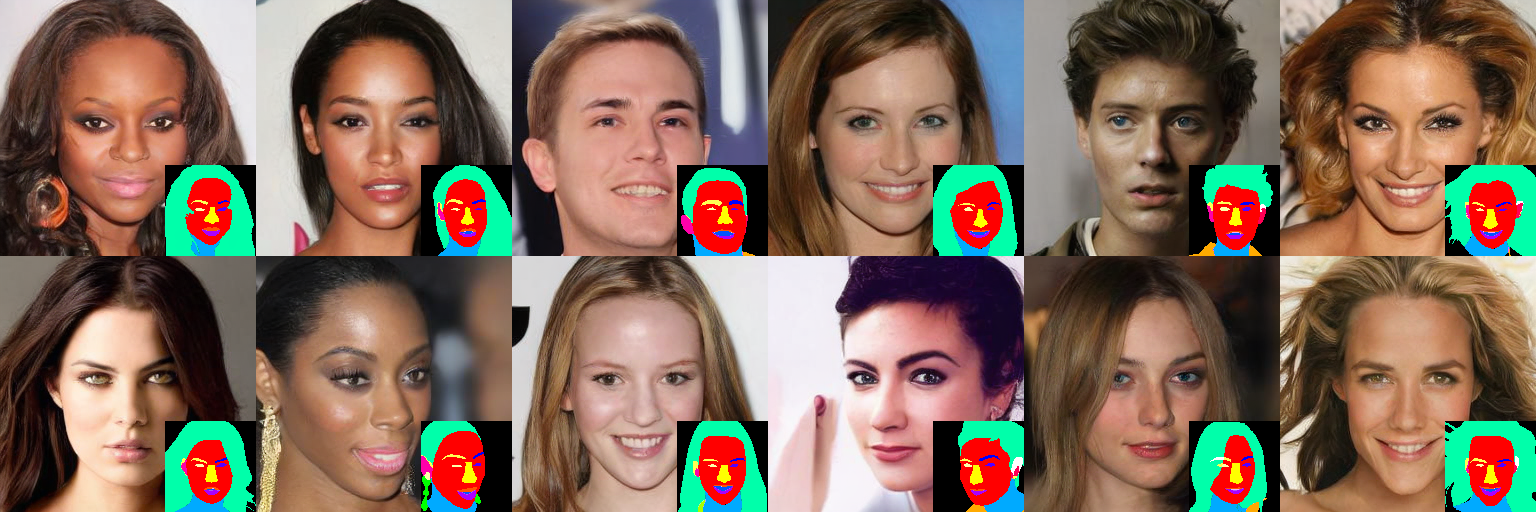}
        \caption{Additional results for Mask-Conditioned Generation.}
        \label{fig:mask}
    \end{figure}

\begin{figure}[!h]
    \centering
        \includegraphics[width=0.9\linewidth]{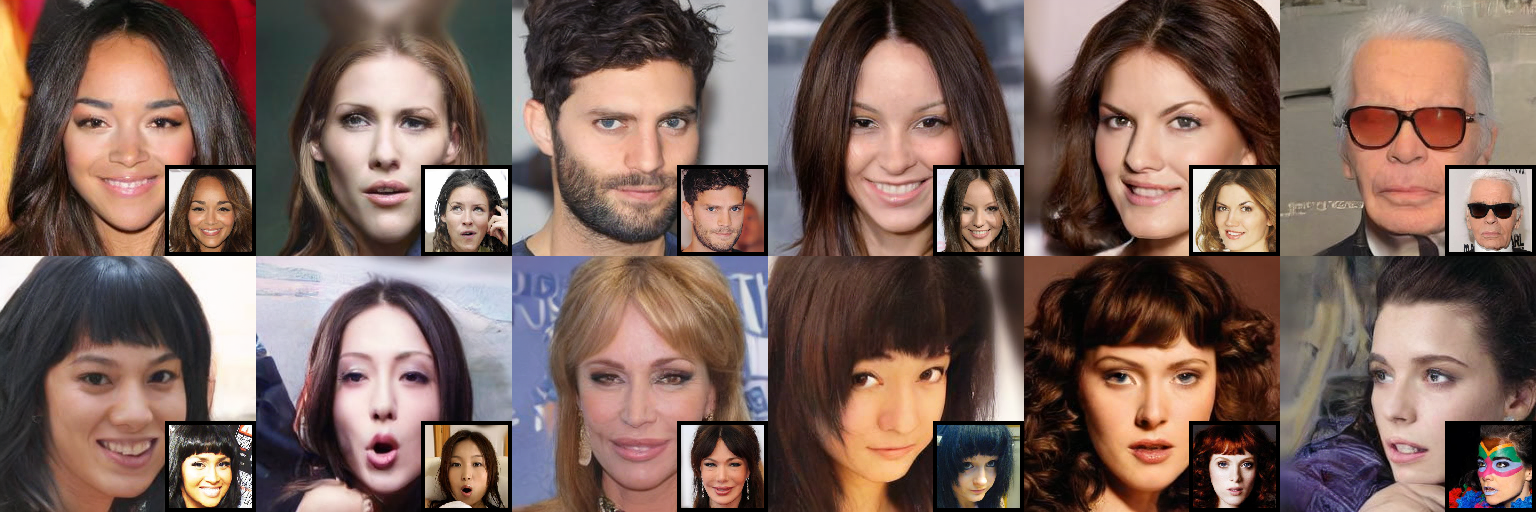}
        \caption{Additional results for Multi-Conditioned Generation.}
        \label{fig:multi}
    \end{figure}

    Figures \ref{fig:attr}, \ref{fig:mask} and \ref{fig:multi} show additional qualitative results on Attributes-Conditioned Generation, Mask-Conditioned Generation, and Multi-Conditioned Generation respectively.
    The description of these experiments can be found in their relative section.
    
\end{document}